\documentclass[sigconf,nonacm]{acmart}
\usepackage{booktabs}
\setcopyright{none}
\renewcommand\footnotetextcopyrightpermission[1]{}

\newcommand{\TCLQueries}{61,763}

\newcommand{\DyGTraining}{22.34}
\newcommand{\DyGHistory}{0.38}
\newcommand{\DyGTrainingSeedOne}{22.27}
\newcommand{\DyGTrainingSeedTwo}{22.42}
\newcommand{\DyGHistorySeedOne}{0.37}
\newcommand{\DyGHistorySeedTwo}{0.39}
\newcommand{\DyGExposure}{87.53}
\newcommand{\DyGFlip}{3.93}

\begin{document}
\title{Interpreting Timestamp Permutations in Dynamic Link Prediction}
\author{Ilia Karpov}
\affiliation{%
  \institution{International Laboratory for Applied Network Research, HSE University}
  \city{Moscow}
  \country{Russia}}
\email{karpovilia@gmail.com}
\author{Kirill Koloskov}
\affiliation{%
  \institution{Independent Researcher}
  \country{Russia}}
\email{koloskov0165@gmail.com}
\author{Valeriy Kuleshov}
\affiliation{%
  \institution{Independent Researcher}
  \country{Russia}}
\email{kuleshov.val@gmail.com}
\author{Dmitry Ilvovsky}
\affiliation{%
  \institution{HSE University}
  \city{Moscow}
  \country{Russia}}
\email{dilv_ru@yahoo.com}

\begin{abstract}
Timestamp permutations are used to test whether dynamic link predictors exploit temporal information, but a performance gap can reflect changes in training, evaluation history, or both. We separate these effects with a crossed diagnostic and measure input exposure: whether altered histories actually change the predictor's inputs. On MOOC, TCL trained with binary cross-entropy gains approximately eight pairwise-accuracy points from original-chronology training under either evaluation history, while changing history at fixed weights gives only 0.03 points. The small history contrast reflects unchanged inputs for 68.37\% of queries, stable rankings for most others, and cancellation of gains and losses. Under original evaluation history, suppressing explicit time encoding reduces the training contrast from 8.03 to 3.64 points, a paired reduction of 4.39 points (95\% interval [3.68, 5.09]) across six seeds. DyGFormer shows corresponding training and history contrasts of \DyGTraining{} and \DyGHistory{} points across two seeds, despite greater exposure. Partial history permutations further separate nominal intervention strength, input changes, and aggregate accuracy. These comparisons turn a single permutation gap into interpretable statements about training conditions and model responses, while leaving the internal mechanisms of temporal learning open.

\end{abstract}
\ccsdesc[500]{Information systems~Data mining}
\ccsdesc[300]{Computing methodologies~Machine learning}
\keywords{temporal graphs, dynamic link prediction, evaluation, timestamp permutation}
\maketitle

\section{Introduction}
A drop in dynamic link-prediction accuracy after permuting timestamps is often interpreted as evidence that a model uses temporal information~\cite{rahman2025counterfactual}. This test reassigns event times while preserving interacting pairs and non-temporal attributes: useful recency or ordering relationships may disappear even though the observed interactions remain. The difficulty is deciding what the resulting gap measures. Retraining on permuted events changes the learning procedure; supplying a changed history to fixed weights changes evaluation inputs. A single comparison that changes both cannot separate their effects.

A small gap is also ambiguous. A predictor that selects recent neighbors may never receive the altered events because later interactions displace them. Even when inputs change, candidate rankings may remain stable, and opposing ranking changes may cancel in aggregate accuracy. Interpretation therefore requires following the intervention from event history through model inputs to predictions.

Existing work tests fixed models on distorted evaluation data~\cite{rahman2025counterfactual}, examines timestamp granularity~\cite{hayes2025learn}, and evaluates causal shifts using known graph generators~\cite{rahman2026causal}. We add a controlled comparison of training and evaluation-history interventions, together with direct checks of the inputs they change. This matters for benchmarking~\cite{tgb2023,yu2023dygformer}: high predictive quality can also arise from edge memorization~\cite{poursafaei2022towards} or recency/popularity heuristics~\cite{cornell2025heuristics}. Our paired experiments across four public datasets show widely varying permutation sensitivity, motivating a closer analysis of what each gap contains.

We ask three questions. How much of a permutation gap comes from training versus evaluation history? When do altered histories change inputs and rankings, and how does intervention strength affect that response? How do explicit time encoding, architecture, and candidate construction shape these conclusions?

Our contribution is a reproducible, crossed, exposure-aware diagnostic that answers these questions through separate contrasts and input tracing. A six-seed study of TCL with binary cross-entropy (TCL-BCE) on the massive open online course (MOOC) dataset~\cite{kumar2019jodie} connects these measurements to a matched time-encoding ablation. DyGFormer and logistic controls test the diagnostic's applicability and distinguish model sensitivity from useful chronology in the prediction task.

\section{Diagnostic Design}
\subsection{Crossing training chronology and history}
The diagnostic varies two factors independently. \emph{Training chronology} is the original or permuted chronology used to learn weights. \emph{Evaluation history} is the original or permuted history used to construct the trained model's inputs. Their four combinations are shown below.

\noindent\begin{minipage}{\columnwidth}
\begin{center}
\begin{tabular}{@{}lcc@{}}
\toprule
 & \multicolumn{2}{c}{Evaluation history}\\
Training chronology & Original & Permuted\\
\midrule
Original & A & B\\
Permuted & C & D\\
\bottomrule
\end{tabular}
\end{center}
\textbf{A} uses original training and original evaluation history. \textbf{B} evaluates the \emph{same original-trained model} with permuted history. \textbf{C} uses permuted training and original evaluation history. \textbf{D} evaluates the \emph{same permutation-trained model} with the corresponding permuted history. All four conditions use identical test queries, query timestamps, and positive/negative candidates. A$\to$B and C$\to$D change only evaluation history, with no retraining.
\end{minipage}

\begin{samepage}
For seed $s$ and permutation $\pi$, let $\theta_{s,0},\theta_{s,\pi}$ be the selected original-trained and permutation-trained weights, and $H_0,H_\pi$ the evaluation histories. For performance measure $M$, the conditions are
\[
\begin{aligned}
 A&=M(\theta_{s,0},H_0), & B&=M(\theta_{s,0},H_\pi),\\
 C&=M(\theta_{s,\pi},H_0), & D&=M(\theta_{s,\pi},H_\pi).
\end{aligned}
\]
\end{samepage}
\textbf{Training-procedure contrasts (A--C and B--D).} A--C compares original- and permutation-trained models under common original history; B--D compares them under the corresponding common permuted history. These contrasts may reflect temporal features, neighborhoods, optimization order, checkpoint selection, or mismatch between training and evaluation histories.

\textbf{A--B and C--D: fixed-weight history contrasts.} A--B measures the effect of changing history for the original-trained model; C--D measures the same intervention for the permutation-trained model. Their difference,
\begin{equation}
 I=(A-B)-(C-D),
\end{equation}
is the \emph{interaction}, also equal to $(A-C)-(B-D)$: does the response to evaluation history depend on training chronology? \textbf{A--D}, the \emph{joint contrast}, changes training chronology and evaluation history simultaneously. The identity $A-D=(A-C)+(C-D)$ splits the joint gap into a training contrast under common history and a fixed-weight history contrast.

We freeze the chronological split and node-exclusion rule. Permutations reassign timestamps within MOOC's 288,224-event pre-split history, preserving endpoints and features. Training uses the same 227,485 eligible events; excluded-node events can still enter history. Histories contain only events strictly preceding the query, and validation/test events retain their original times.

\begin{table*}[t]
\centering\small
\caption{Crossed MOOC diagnostic. A--D: pairwise accuracies (\%); contrasts: percentage points, computed before rounding. TCL-BCE averages five permutations within each seed, then six seeds. DyGFormer reports seed means. ON/OFF: explicit time encoding during training and inference. AP, centered AP, individual permutations, and intervals are in the artifact.}
\label{tab:crossed}
\begin{tabular}{@{}llrrrrrrrrr@{}}
\toprule
Configuration & Seeds & A & B & C & D & A--C & B--D & A--B & C--D & $I$\\
\midrule
TCL-BCE ON & 6 & 81.90 & 81.87 & 73.87 & 73.92 & 8.03 & 7.95 & 0.03 & -0.04 & 0.07\\
TCL-BCE OFF & 6 & 70.00 & 69.92 & 66.36 & 66.39 & 3.64 & 3.53 & 0.08 & -0.03 & 0.11\\
\midrule
DyGFormer & 1004 & 86.90 & 86.54 & 64.64 & 64.37 & 22.27 & 22.17 & 0.37 & 0.27 & 0.10\\
DyGFormer & 1005 & 86.95 & 86.57 & 64.54 & 64.29 & 22.42 & 22.27 & 0.39 & 0.24 & 0.14\\
\bottomrule
\end{tabular}

\end{table*}

\subsection{Input exposure and prediction quality}
TCL-BCE uses only the 20 most recent interactions for each endpoint. Later validation and test events can push altered training events outside this window, leaving a query's inputs unchanged. For these queries, A--B cannot test the response to altered temporal information.

For each query, compare the tensors supplied under $H_0$ and $H_\pi$ for the source and both candidates. For TCL-BCE these include ordered node/edge features, time differences, and padding masks. \emph{Input exposure} is the fraction of queries for which any compared input changes. The unchanged-input group has identical inputs across all tested histories. At fixed weights, equal inputs imply equal predictions up to numerical precision. Graded exposure $d_q$ divides changed history positions by positions nonempty in either history, summed over source and both candidates (zero if empty). A position changes when its features, time difference, or padding changes. 

Pairwise accuracy credits one when the positive candidate outranks the negative and one half for a tie. Pooled average precision (AP) also ranks examples across queries. To diagnose shared query offsets, subtract each query's mean logit from both logits and recompute AP. This label-free centering preserves within-query rankings but does not decompose AP additively. Accuracy and AP contrasts use percentage points; absolute AP uses the 0--1 scale.

\section{Experiments and Findings}
\subsection{Controlled TCL-BCE evaluation}
We evaluate DyGLib-based TCL-BCE~\cite{wang2021tcl,yu2023dygformer} on MOOC learner--resource interactions~\cite{kumar2019jodie}. All \TCLQueries{} test queries have one fixed random negative, including 670 positive--negative collisions. Training negatives are keyed by event ID, seed, and epoch over a fixed destination pool, so they match across chronologies for each event and shared epoch.

For each of six seeds (1004--1009), we train on original chronology and five permutations. Checkpoints use the earliest strict maximum of pooled validation AP, with 20--100 epochs and patience 10. We average permutation contrasts within seed; descriptive 95\% $t_5$ intervals use six seed means as uncertainty units.

All experimental code and trained models are available for download from the project repository at \url{https://github.com/karpovilia/tgn}.

\subsection{Separating training and history effects}
Original training improves pairwise accuracy under both common evaluation histories: by 8.03 points with original history (A--C; 95\% interval [7.65, 8.41]) and 7.95 with permuted history (B--D; \mbox{[7.59, 8.31]}). Both are positive in every seed. Changing only the original-trained model's evaluation history gives a 0.03-point contrast (A--B). The joint A--D gap of 7.98 points mainly reflects the training contrast. The interaction is 0.074 points, with a 95\% interval of \mbox{[0.023, 0.125]}. The A--C advantage is also positive in every seed for pooled AP (13.50 points) and centered AP (9.71 points; intervals in the artifact). Excluding candidate collisions retains an 8.11-point accuracy contrast [7.73, 8.50] and 13.83 AP points over 61,093 queries.

To understand the small history contrast, we first check whether the intervention reaches the model. Each permutation changes inputs for the same 19,533 queries (31.63\%); the other 42,230 (68.37\%) retain identical inputs. Exposure falls from 39.15\% in the first equal-count test decile to 14.04\% in the last. 

\begin{figure*}[t]
\centering
\includegraphics[width=\textwidth]{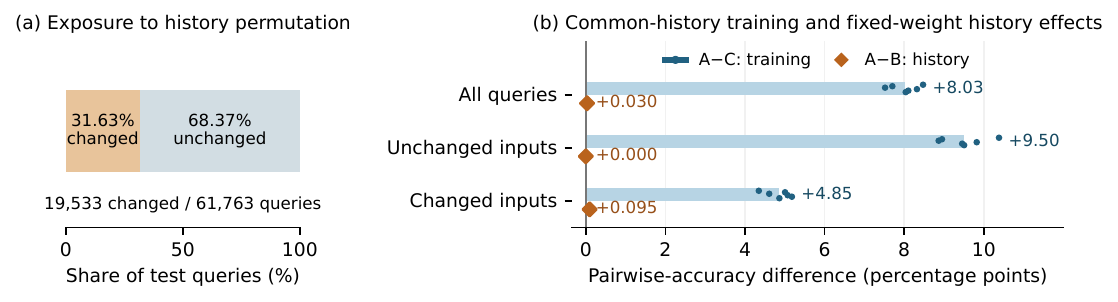}
\caption{TCL-BCE input exposure and accuracy contrasts. (a) Queries with changed or identical inputs. (b) A--C: blue means and six seed dots; A--B: orange diamonds at fixed original-trained weights. Contrasts average five permutations within seed and are in percentage points. Input-defined groups differ in composition.}
\Description{31.63 percent of queries have changed inputs. Training contrasts are 8.03 points overall, 9.50 for unchanged inputs and 4.85 for changed inputs; history contrasts are about 0.03, zero and 0.095 points.}
\label{fig:diagnostic}
\end{figure*}

Exposure does not guarantee a ranking change. Only 2.77\% of exposed queries change ordering (0.875\% of all queries), and opposing gains and losses leave an A--B contrast of 0.095 points; C--D is $-0.139$ points in this group. Unchanged inputs produce no history-induced ranking flips in either training condition. Their maximum logit drift is $6.63\times10^{-6}$ across 60 comparisons, within the prespecified tolerance $2\times10^{-5}+10^{-5}(|z_0|+|z_\pi|)$. The small contrast combines unchanged inputs, stable rankings, and cancellation.

A--C is positive in every seed for unchanged-input queries (9.50 points) and exposed queries (4.85 points; Figure~\ref{fig:diagnostic}). The training advantage persists where history interventions leave inputs untouched. These groups differ in composition.

\paragraph{Graded exposure.}
Input changes also vary in intensity. With original-trained weights, ranking flips rise from 2.19\% [1.91, 2.47] for $0<d_q\leq0.25$ to 4.77\% [4.05, 5.49] for $0.25<d_q\leq0.5$; mean absolute changes in the positive-minus-negative logit margin rise from 0.197 to 0.458. Intervals use six seed means, each averaging five permutations. About 15,160 and 4,343 queries per permutation fall in these groups; only 30 have $d_q>0.5$. Differing temporal composition makes this comparison descriptive.

We then vary intervention strength while holding weights fixed, permuting timestamps within nested random subsets of the history prefix, with fractions $\alpha\in\{0,.25,.5,.75,1\}$. Selected events are ordered by their positions in the full permutation and assigned, in that order, to their selected original timestamp slots in chronological order. Unselected events keep their timestamps; $\alpha=0$ is identity and $\alpha=1$ is the full permutation.

\begin{samepage}
Both original-trained and full-permutation-trained weights stay fixed across the six seeds and five permutations. From $\alpha=.25$ to $1$, mean exposure increases from 27.91\% to 31.63\%, while all-query ranking flips increase from 0.585\% to 0.875\% at original-trained weights; the corresponding accuracy contrasts remain 0.022--0.033 points across nonzero levels. Thus a stronger event-level intervention can change more rankings while leaving aggregate accuracy nearly unchanged. History construction also matters: exposure is 12.01\%, 18.54\%, 31.63\%, and 62.64\% at 5, 10, 20, and 50 neighbors. This checks which inputs are reached, without comparing model quality across history lengths.
\par\end{samepage}

\subsection{Explicit time encoding}
To test how the training advantage depends on explicit time encoding, we repeat A--D with zero time-encoder outputs during training and inference. Module layout, architecture, neighborhood construction, timestamp sorting, candidates, and checkpoint selection are preserved. Nine OFF models from the three-seed, two-permutation validation pilot are reused.

With encoding OFF, A--C falls from 8.03 to 3.64 points [2.91, 4.37] (Table~\ref{tab:crossed}). Define
\[
\Delta_{\rm time}=(A-C)_{\rm ON}-(A-C)_{\rm OFF}.
\]
The paired reduction is 4.39 points [3.68, 5.09], positive for all six seeds. Removing candidate collisions gives a 4.43-point reduction [3.72, 5.15]. AP and centered AP give positive paired reductions of 7.49 and 4.97 points. Removing explicit time encoding reduces the measured contrast by 54.7\%. Because the ablation changes the fitted predictor, this is component sensitivity, not a causal decomposition of learned temporal information. The remaining contrast may involve temporal neighborhoods and optimization order.

\subsection{Transfer to DyGFormer}
To test applicability beyond TCL-BCE, we apply A--D to DyGFormer~\cite{yu2023dygformer} on the same split, queries, and candidates. Seeds 1004 and 1005 form a hardware-defined block independent of test effects, each with five permutations. Training uses BCE, 256-event sequences, and TCL-BCE's selection and epoch rules. Negatives follow the same event-ID/seed/epoch schedule and fixed destination pool across chronologies. A two-permutation validation pilot preceded test evaluation. We report seed means separately because permutations sharing a seed are dependent.

The A--C accuracy contrasts are \DyGTrainingSeedOne{} and \DyGTrainingSeedTwo{} points; A--B is \DyGHistorySeedOne{} and \DyGHistorySeedTwo{} points (Table~\ref{tab:crossed}). DyGFormer's input trace includes node/edge/time features, padding, and neighbor co-occurrence counts; hashes are compared only at fixed weights. The history intervention exposes \DyGExposure{}\% of queries, compared with 31.63\% for TCL-BCE, and changes candidate ordering for \DyGFlip{}\% of exposed queries (3.44\% of all queries) under original-trained weights. The 256-event DyGFormer sequences and 20-interaction TCL-BCE histories differ in available context, so exposure differences concern these configurations. Equal inputs preserve rankings in every checked comparison. Excluding candidate collisions retains the positive training-procedure contrast in both seeds. The protocol therefore transfers across architectures: even with greater exposure, stable rankings and cancellation keep the history contrast much smaller than the training contrast.

\subsection{Recency and multi-dataset comparisons}
The neural contrasts describe model responses; logistic controls ask whether chronology supplies useful predictive features under matched candidates. Full-batch logistic regression compares a six-feature baseline with the same features plus ages since last occurrence. The six baseline features are pair, source, and destination interaction counts, with three corresponding seen indicators. Counts and ages are log-transformed, ages use a TRAIN-derived time scale, and standardization uses TRAIN only. Feature and optimizer specifications are in the artifact.

Original/permuted fits share training queries, row order, negative identities, and original causal evaluation history. Historical TRAIN negatives are destinations previously observed with the source in all compared histories; random negatives use the fixed TRAIN destination universe. Both exclude the positive candidate. Queries lacking a common historical alternative are excluded from both protocols but retained as history events. At test time, historical negatives precede the query; random negatives use the same universe. Both protocols evaluate 60,630 MOOC and 14,515 Wikipedia queries after exclusions and collision removal. Each chronology comparison fixes candidates within its protocol; absolute accuracy across these different prediction tasks is not a measure of model superiority.

Let $\Delta_{\rm ages}$ and $\Delta_{\rm counts}$ be original-minus-global-permutation accuracy contrasts, and $\Delta_{\rm recency}=\Delta_{\rm ages}-\Delta_{\rm counts}$. Across ten paired draws, historical negatives yield positive recency contrasts on MOOC (33.90 points [33.18, 34.61]) and Wikipedia (16.15 [15.41, 16.89]). With random TRAIN/test negatives, MOOC retains a 14.17-point contrast [13.99, 14.35], positive in every draw (Table~\ref{tab:candidates}). Recency utility therefore survives this change of candidate distribution, although its magnitude changes. A three-draw validation pilot preceded the frozen random-protocol evaluation. This MOOC population passes the recorded composition checks: less than 95\% unseen negatives, at least 1,000 both-seen queries, and accuracy below 95\% for a rule using pair-seen status alone, in every draw. On Wikipedia, random negatives yield a 1.63-point recency contrast [0.99, 2.28]. In every draw, unseen negatives exceed 95\% and fewer than 1,000 queries have both candidates seen. We therefore restrict the random-sampling robustness claim to MOOC. Wikipedia results and windowed controls are reported in the artifact.

\begin{table}[htbp]
\centering\small
\caption{MOOC logistic training-procedure contrasts (points), with 95\% paired $t_9$ intervals over ten draws. H/R: historical/random negatives in both TRAIN and test. Queries are common across protocols; candidates are fixed within each chronology contrast.}
\label{tab:candidates}
\begin{tabular}{@{}lrr@{}}
\toprule
Dataset / negatives & $\Delta_{\rm ages}$ & $\Delta_{\rm recency}$\\
\midrule
MOOC / H & 32.67 [32.14, 33.20] & 33.90 [33.18, 34.61]\\
MOOC / R & 17.21 [17.06, 17.36] & 14.17 [13.99, 14.35]\\
\bottomrule
\end{tabular}

\end{table}

\paragraph{Multi-dataset permutation experiments.}
Our 195 DyGLib original/permuted pairs cover five architectures and Wikipedia, Reddit, MOOC, and LastFM, using ten-epoch budgets, random negatives, one training-time permutation, and joint validation AP/AUC selection.

\begin{samepage}
TCL-BCE original-minus-permuted AP gaps are 5.55, $-0.12$, 10.76, and 24.23 points on these datasets, respectively (nine unique pairs each).
\par\end{samepage}
On LastFM, DyGFormer achieves original AP 0.9245 with a 0.19-point gap, compared with AP 0.7552 and a 24.23-point gap for TCL-BCE (ten and nine pairs). Across all 20 model--dataset combinations, mean AP gaps range from $-2.04$ to 24.23 points: five are negative and ten have magnitude below 0.20 points. The four-dataset comparisons characterize variation in permutation sensitivity; the controlled MOOC experiments separate training-procedure and fixed-weight evaluation-history contrasts.

\section{Interpretation and Scope}
A--C and B--D measure the original training-procedure advantage under common original and permuted histories, respectively. The ON/OFF ablation measures its dependence on explicit time encoding. A--B and C--D measure the response of fixed weights to altered history; exposure identifies which inputs can contribute, and ranking changes show how individual responses combine in accuracy. Graded and partial permutations connect intervention strength to these responses. These measurements explain why a small fixed-weight history contrast can coexist with a large training-procedure contrast.

Logistic controls address useful recency in a candidate-defined prediction task. The four-dataset comparisons describe variation across models and data. Neither substitutes for the crossed contrasts, and the component ablation does not identify a unique internal temporal-learning mechanism. Conclusions concern the evaluated configurations and candidate populations. Test data were previously examined, and pilots used the checkpoint-selection split. Intervals describe fixed-dataset seed/draw variability; DyGFormer reports two seed means.

\section{Conclusion}
Training chronology and evaluation history require separate analysis. For TCL-BCE, a large training-procedure advantage coexists with a small fixed-weight history contrast because many inputs remain unchanged, most exposed rankings are stable, and opposing changes cancel. Graded and partial permutations show how intervention strength reaches predictions, while time-encoding ablation reduces the training contrast without eliminating it. DyGFormer confirms the diagnostic's applicability under a different exposure profile. Permutation studies should therefore report training and history contrasts, actual input exposure, and ranking changes, with uncertainty aggregated within training seeds.

\label{sec:mainend}
\section*{Ethical Considerations}
This study reuses public interaction benchmarks and reports aggregate model behavior. Reproducibility materials preserve pseudonymous identifiers and should respect source-data licenses without releasing additional identifying information. No individual-level behavioral claims are made. The diagnostic distinguishes measured protocol effects from claims about internal mechanisms or deployment performance. Generative artificial intelligence assisted code development and manuscript editing, including improvements to language and presentation. The authors take full responsibility for all content, including verifying its correctness and originality and ensuring proper attribution. Reported quantities were checked against experimental artifacts.
\balance
\bibliographystyle{ACM-Reference-Format}
\bibliography{references}
\end{document}